\documentclass[10pt,twocolumn,letterpaper]{article}

\usepackage{cv}              

\definecolor{iccvblue}{rgb}{0.21,0.49,0.74}
\usepackage[pagebackref,breaklinks,colorlinks,allcolors=iccvblue]{hyperref}
\usepackage{multirow}

\title{Stealthy Patch-Wise Backdoor Attack in 3D Point Cloud via Curvature Awareness}

\author{Yu Feng\textsuperscript{1} \quad
Dingxin Zhang\textsuperscript{1}\quad
Runkai Zhao\textsuperscript{1}\quad
Yong Xia\textsuperscript{2}\quad
Heng Huang\textsuperscript{3}\quad
Weidong Cai\textsuperscript{1}\\
$^1$ The University of Sydney\\
$^2$ Northwestern Polytechnical University \quad $^3$ University of Maryland College Park\\
{\tt\small  \{yfen0146,dzha2344\} @uni.sydney.edu.au} \quad \tt\small runkai.zhao@sydney.edu.au\\ 
 {\tt\small  yxia@nwpu.edu.cn\quad heng@umd.edu\quad tom.cai@sydney.edu.au}
}

\begin{document}
\maketitle

\begin{abstract}
Backdoor attacks pose a severe threat to deep neural networks (DNNs) by implanting hidden backdoors that can be activated with predefined triggers to manipulate model behaviors maliciously. 
Recent studies have extended backdoor attacks to 3D point clouds, but most existing triggers are sample-wise and often cause visible geometric artifacts or high optimization cost. 
To address these limitations, we propose the \textbf{S}tealthy \textbf{P}atch-Wise \textbf{B}ackdoor \textbf{A}ttack (\textbf{SPBA}), a patch-wise backdoor attack framework for 3D point clouds.
Specifically, SPBA decomposes a point cloud into local patches, where each patch is formed by a Farthest Point Sampling (FPS) center and its K-nearest neighbors (KNN). Candidate patches are ranked using a patch imperceptibility score derived from local curvature variation, and a unified spectral trigger is injected into the selected patches by perturbing only the coordinates of existing points while preserving the original point cardinality.
Extensive experiments on ModelNet40 and ShapeNetPart further demonstrate that SPBA achieves the state-of-the-art stealthiness among the prior methods and reduces spectral-trigger computation by 98.43\% relative to a sample-wise spectral baseline, while maintaining competitive attack performance. 
These results support localized spectral design as an effective and efficient approach to stealthy backdoor attacks in 3D point cloud models.
Code is available at \url{https://github.com/HazardFY/SPBA}.
\end{abstract}

\section{Introduction}

Point clouds serve as a highly effective data structure for representing 3D geometric data~\cite{guo2020pointcloud}, finding increasing utility in safety-critical applications including autonomous driving~\cite{li2020driving}, healthcare~\cite{yu20213d}, and robotics~\cite{zhu2025point}.  With this expanding deployment, the security of point cloud models becomes paramount, particularly concerning their vulnerability to backdoor attacks~\cite{li2021pointba,peng2025backdoor}. 
In these attacks, adversaries inject a pre-defined trigger into a small portion of training data. This manipulation causes the backdoored model, upon training, to consistently misclassify samples containing the trigger into a target class, while maintaining accurate predictions for benign inputs.
A well-designed trigger remains inconspicuous within the data yet potently activates malicious behavior in the backdoored model. Therefore, trigger design is crucial for both the effectiveness and stealth of backdoor attacks~\cite{huang2026mfbd,li2025dtgba}.

\begin{figure}[!t]
    \centering
    \includegraphics[width=1.0\linewidth]{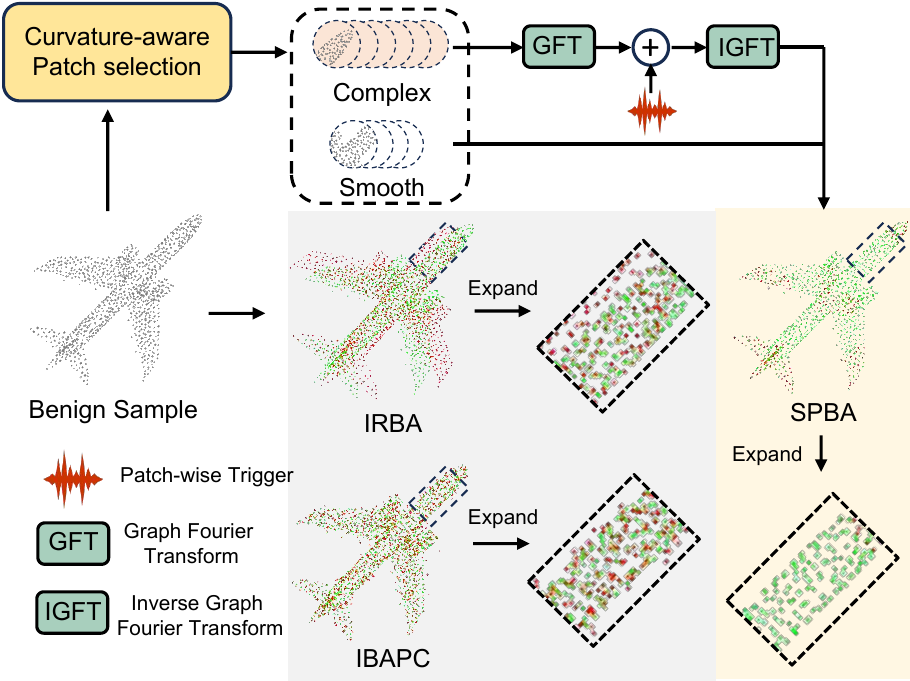}
    \caption{Comparison of latest sample-wise attacks and our stealthy patch-wise attack. Green points represent the original sample, while red points indicate modifications introduced by injected triggers.}
    \label{fig:begin}
\end{figure}

Current 3D point cloud backdoor attacks are largely dominated by sample-wise trigger injection. 
Early explorations into 3D point cloud backdoor attacks utilized direct global geometric modifications as triggers, such as adding spherical point clusters or applying specific rotations~\cite{li2021pointba,xiang2021backdoor,yuan2023transformer}.
These global modifications produce deformations readily discernible to the human eye and susceptible to standard data preprocessing techniques like Statistical Outlier Removal (SOR)~\cite{zhou2019dup} or rotation augmentation~\cite{gao2023IRBA}.
Recent efforts have shifted towards incorporating local context into trigger design to enhance stealthiness. 
For instance, IRBA~\cite{gao2023IRBA} utilized a nonlinear and local transformation to generate sample-wise triggers, while IBAPC~\cite{fan2024IBAPC} employed global spectral triggers to balance global shape preservation with localized point perturbations.
These triggers are more adaptable, fitting the local geometric structure of samples while preserving the overall global geometry for enhanced stealthiness. 
However, these methods do not sufficiently focus on local context, limiting their potential for improved stealthiness. 
Their triggers either alter local-to-global spatial relationships~\cite{gao2023IRBA} or rely on global spectral optimization with relatively high computational cost~\cite{fan2024IBAPC}.

These limitations motivate our study of patch-wise trigger design. Such a design is straightforward in 2D images, where patches are naturally defined by the image grid~\cite{gu2019badnets}. In point clouds, however, patches are not native primitives because point sets are unordered, sparse, and irregular. As a result, an effective patch-wise backdoor for 3D point clouds requires a geometry-aware patch definition, a transferable trigger parameterization that does not depend on point indices, and a reconstruction mechanism that preserves point-set cardinality while suppressing geometric artifacts.
To satisfy these requirements, we adopt a localized spectral formulation. Instead of optimizing perturbations directly in the raw spatial domain, the trigger is parameterized over local patch representations in the spectral domain, which provides more structured control over local deformation. In addition, prior work suggests that perturbations in geometrically complex regions tend to be less perceptible than those in smoother regions~\cite{lou2024hide}. This observation motivates geometry-aware patch selection for improving stealthiness, as illustrated in Fig.~\ref{fig:begin}. Compared with sample-wise spectral optimization, such a patch-wise formulation is also more computationally efficient.

Based on this design, we propose \emph{Stealthy Patch-Wise Backdoor Attack} (SPBA), a patch-wise spectral backdoor attack for 3D point clouds. SPBA first decomposes each point cloud into local patches using FPS anchors and KNN neighborhoods, enabling fine-grained control over trigger placement without relying on point ordering. To identify suitable trigger regions, we introduce a curvature-based Patch Imperceptibility Score (PIS) that quantifies local geometric complexity, and apply the trigger only to selected high-PIS patches. The selected patches are represented as local graphs, whose coordinates are transformed into the frequency domain via the Graph Fourier Transform (GFT). A unified spectral trigger is then optimized on these patches and mapped back to the original point coordinates through the inverse transform. Because SPBA perturbs only the coordinates of existing points within selected patches, it better preserves the overall shape and avoids conspicuous distortions. The poisoned patches are finally merged with the unmodified regions to form the poisoned point cloud. During training, the trigger and the backdoored model are jointly optimized through an alternating procedure to balance attack effectiveness and stealthiness.

Our main contributions include:

\begin{itemize}

\item We propose SPBA, a patch-wise backdoor attack on 3D point clouds that leverages the Graph Fourier Transform (GFT) to enable efficient and precise spectral perturbations at the local level while preserving the global geometric structure.

\item  We introduce a curvature-based patch imperceptibility scoring mechanism to identify and select geometrically complex patches for trigger injection, enhancing the stealthiness and effectiveness of SPBA.

\item Comprehensive experiments conducted on two public benchmarks to demonstrate that SPBA achieves the best stealthiness among the compared methods, maintains competitive attack performance, and significantly improves efficiency relative to a sample-wise spectral method.
\end{itemize}

The rest of this paper is organized as follows. Section 2 reviews patch-wise backdoor attacks in 2D images and summarizes prior backdoor attacks and defenses in 3D point clouds. Section 3 introduces the preliminaries of the proposed SPBA framework. Section 4 presents the proposed SPBA method in detail. Section 5 describes the datasets and experimental settings, followed by the experimental results and discussion. Section 6 discusses the implications and limitations of SPBA, and outlines several directions for future work. Finally, Section 7 concludes the paper.

\section{Related Work}
\subsection{Patch-Wise Backdoor Attacks in 2D Images}
Different from adversarial attacks, backdoor attacks pose a significant security threat in the training process by embedding hidden triggers during the training phase, enabling adversaries to manipulate model predictions during inference~\cite{song2026wpda, yang2026featuretrojan}. 
Patch-wise triggers are a well-established design in the image domain, as the regular pixel grid naturally defines local regions for perturbation.
BadNets first showed that a small fixed image patch can reliably activate a target label after poisoning~\cite{gu2019badnets}. 
Blended attacks~\cite{chen2017blended} then improved visual imperceptibility by mixing the trigger with the input image, and later sparse or optimized patch attacks further improved the balance between attack success and visual detectability~\cite{gao2024backdoor,Yang_2024_CVPR}.
The patch-wise idea has also been studied in more recent learning paradigms. Patch-style backdoors can survive self-supervised pre-training~\cite{saha2022backdoor}, compromise multimodal contrastive alignment~\cite{Bai_2024_BadClipOp}, and exploit the patch-token structure of Vision Transformers~\cite{yuan2023transformer}. Patch triggers have also been used for benign purposes such as dataset ownership verification~\cite{li2022untargeted}. 
These studies suggest that patch-wise triggers could be both effective and imperceptible. However, image patches rely on an ordered grid and fixed spatial neighborhoods. This assumption does not hold for point clouds, where a local trigger must be defined over unordered 3D coordinates while preserving point cardinality and local surface geometry.

\subsection{Backdoor Attacks and Defenses in 3D Point Clouds}
Existing 3D point cloud backdoor attacks can be grouped by the type of trigger they use. Early methods introduce direct geometric changes, such as adding point clusters or applying predefined rotations~\cite{li2021pointba,xiang2021backdoor}. These attacks are easy to implement, but inserted point clusters can be visually noticeable and sensitive to outlier removal, while rotation triggers can be weakened by rotation augmentation.
More recent 3D backdoor attacks make the trigger better match the point-cloud geometry. IRBA~\cite{gao2023IRBA} uses nonlinear local transformations to generate sample-specific triggers, while IBAPC~\cite{fan2024IBAPC} formulates a sample-wise spectral trigger to preserve global geometric structure. Other designs introduce perturbations through reconstruction~\cite{bian2024iba}. These methods show that geometry-aware trigger design can improve stealthiness, but sample-wise optimization and reconstruction modules can reduce efficiency and generality.
A key distinction from prior 3D backdoor attacks is that SPBA adopts a patch-wise trigger design rather than sample-wise optimization or direct point insertion. Restricting perturbations to geometrically defined local patches leads to a more favorable balance between stealthiness and efficiency by better preserving the overall shape and reducing conspicuous local artifacts.


On the defense side, conventional data augmentations are often employed as a defense strategy against backdoor attacks~\cite{gao2023IRBA}.
SOR~\cite{zhou2019dup} enhanced robustness by removing outliers from point clouds.
Reverse-engineering defenses attempt to detect backdoored point-cloud classifiers by recovering suspicious trigger patterns~\cite{xiang2022detecting}. PointCRT detects poisoned samples by analyzing abnormal corruption robustness~\cite{hu2023pointcrt}. Cross-modal object restoration removes implanted triggers by reconstructing point clouds from cross-modal cues~\cite{lian2025cross}.  

Additionally, prior point-cloud adversarial attack studies motivate the use of spectral representations and local geometry~\cite{hu2022exploringdevil,liu2023meet}, but they focus on sample-specific test-time perturbations rather than train-time backdoor trigger design. In contrast, SPBA learns a unified trigger that can be injected across samples while retaining local patch-wise support.

\begin{figure*}[t]
    \centering
    \includegraphics[width=1.0\linewidth]{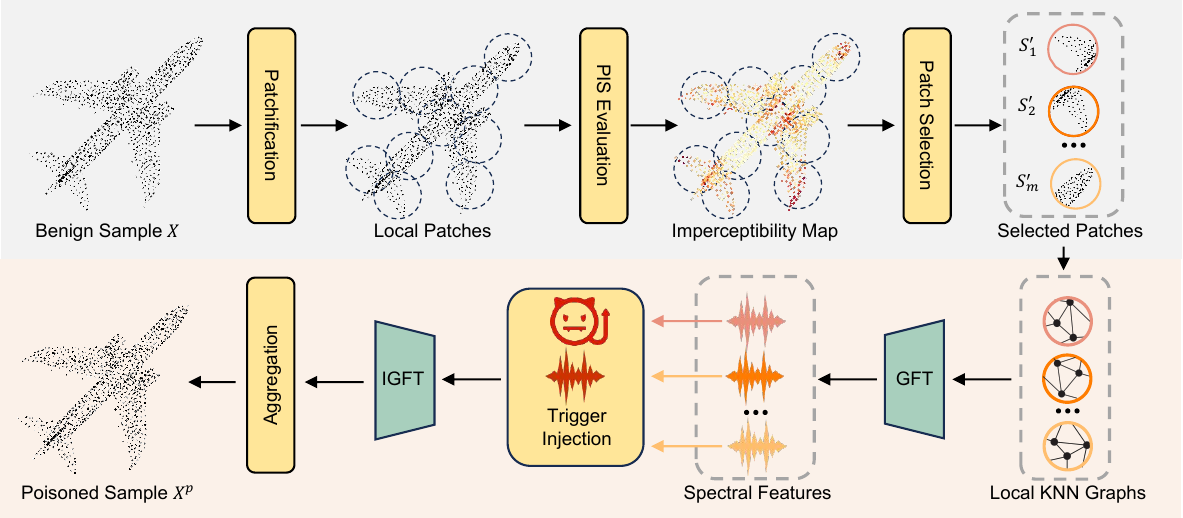}
    \caption{The framework of our proposed SPBA method. Given a benign point cloud sample $X$, SPBA first decomposes the sample into local patches and evaluates their patch imperceptibility scores (PIS).  Patches with high PIS are then selected and transformed into the spectral domain using the Graph Fourier Transform (GFT). A spectral trigger is injected into the selected patches, ensuring stealthy perturbations.  Finally, the Inverse Graph Fourier Transform (IGFT) reconstructs the local patches, which are then aggregated with the unmodified patches to form the final poisoned sample $X^p$.}
    \label{fig:Framework}
\end{figure*}

\section{Preliminaries}
\subsection{Problem Statement}
The training set is defined as $D = \{(X_i, Y_i)\}_{i=1}^N$, where $X_i \in \mathbb{R}^{n \times 3}$ contains the coordinates of the $n$ points in the $i$-th sample and $Y_i \in \{1, \dots, K\}$ is the corresponding class label. Following the standard targeted backdoor setting~\cite{gu2019badnets,feng2022fiba}, $D$ is partitioned into a benign subset $D_b$ and a poisoned subset $D_p$. A fraction $\rho$ of training samples is transformed by a trigger injection function $\mathcal{T}$ and relabeled to a predefined target class $Y_t$. Training a model $f_{\theta}$ on the resulting dataset should preserve high accuracy on benign inputs while causing poisoned inputs $X_i^p = \mathcal{T}(X_i)$ to be classified as $Y_t$.

\subsection{Threat Model}
Consistent with prior backdoor attack methods \cite{xiang2021backdoor,fan2024IBAPC}, we assume that adversaries operate under a white-box setting, where they have full access to the model architecture, can manipulate a subset of the training data, and control the training process. By injecting poisoned samples into the training set, adversaries can implant backdoors into DNNs. At the inference stage, the adversaries only need black-box query access to the trained model.

\section{Method}
\subsection{Patchification and Curvature-Aware Selection}
Due to the unordered and irregular nature of point clouds, it is infeasible to directly segment patches for trigger injection in the same manner as in 2D images. 
Inspired by prior work~\cite{zhang2022pointmae}, we define a patch in point clouds as a spatial neighborhood.
Given an input point cloud $X=\{x_j\}_{j=1}^{n}$, Farthest Point Sampling (FPS)~\cite{zhang2022pointmae} is first used to select $g$ anchor points
$X^{ct}=\{x^{ct}_1,x^{ct}_2,\dots,x^{ct}_{g}\}$.
For each anchor $x^{ct}_i$, K-nearest neighbors (KNN) returns an index set $I_i \subseteq \{1,\dots,n\}$ with $|I_i|=k_g$, and the corresponding local patch is constructed as
\begin{equation}
S_i=X[I_i]=\mathit{K\!N\!N}(X, x_i^{ct}), \quad S_i \in \mathbb{R}^{k_g \times 3}.
\end{equation}
The resulting patch set is denoted by $\{S_1,S_2,\dots,S_g\}$.

The candidate patches are then ranked according to local geometric complexity. 
As human visual sensitivity varies across different regions of a 3D object, perturbations in high-complexity patches are less perceptible due to the absence of a fixed prior shape. 
We define a patch imperceptibility score (PIS), which draws from the imperceptibility score concept of the saliency and imperceptibility score (SI score)~\cite{lou2024hide}. 
Specifically, the Patch Imperceptibility Score (PIS) is defined from the pointwise imperceptibility score introduced below.
The local curvature around each point  $x_j$ in a point cloud $X$ is defined as:
\begin{equation}
C(x_j;X) = \frac{1}{k_c} \sum_{q \in \mathcal{N}_{x_j}} | \langle \frac{q - x_j}{ \|q - x_j\|_2}, \mathbf{n_{x_j}} \rangle |,
\label{eq:curvature}
\end{equation}
where $\mathcal{N}_{x_j}$ is defined as the set of $k_c$ nearest neighbors of point $x_j$ and $\mathbf{n_{x_j}}$ denotes its normal vector.
The imperceptibility score (IS) of each point is further computed as the standard deviation of curvature between the point $x_j$ and its neighboring points $\mathcal{N}_{x_j}$ (with neighborhood size fixed as $k_c=10$, following prior work~\cite{lou2024hide}), capturing local curvature fluctuations:
\begin{equation}
IS(x_j;X) = \sqrt{\frac{1}{k_c} \sum_{q \in \mathcal{N}_{x_j}} (C(q) - \frac{1}{k_c} \sum_{q \in \mathcal{N}_{x_j}} C(q))^2}.
\end{equation}
This measure yields an imperceptibility map for the entire point cloud. 
Finally, the patch imperceptibility score (PIS) of the decomposed local patch $S_i$ is calculated by averaging the score of all points within the patch:
\begin{equation}
\mathit{PIS}(S_i) = \frac{1}{k_g} \sum_{x \in {S_i}} \mathit{IS}(x).
\end{equation}
The $m$ patches with the highest PIS values, denoted by $\{S'_1, S'_2,\dots,S'_m\}$ in descending PIS order with index sets $\{I'_1,I'_2,\dots,I'_m\}$, are selected for subsequent trigger injection to favor less perceptible perturbations.
All PIS values and selected patches are computed on the clean point cloud $X$ before any trigger perturbation is applied, as shown in Fig.~\ref{fig:Framework}.

\subsection{Patch-wise Spectral Trigger Injection}
\label{sec:trigger}

Directly applying a unified spatial trigger across different local regions of 3D point clouds is challenging because their unordered and irregular structure limits the transferability of spatial triggers across samples, often causing noticeable outliers and semantic distortions that undermine both stealthiness and attack effectiveness.
We therefore propose designing local patch-wise triggers in the spectral domain.

For each selected patch $S' \in \{S'_1,S'_2,\dots,S'_m\}$, an unweighted KNN graph $G = (V, A)$ is constructed, where $V \in \mathbb{R}^{k_g \times 3}$ contains the patch coordinates and $A \in \mathbb{R}^{k_g \times k_g}$ connects each point to its $k_p=10$ nearest neighbors~\cite{fan2024IBAPC,hu2022exploringdevil,liu2023meet}. The combinatorial graph Laplacian is then computed as
\begin{equation}
L = D - A,
\end{equation}
where $D \in \mathbb{R}^{k_g \times k_g}$ is the degree matrix. Since $L$ is real, symmetric, and positive semi-definite, it admits the eigen decomposition $L = U \Lambda U^T$, where $U \in \mathbb{R}^{k_g \times k_g}$ contains the orthonormal eigenvectors and $\Lambda$ contains the eigenvalues.

The Graph Fourier Transform (GFT) maps the patch coordinates into the graph-frequency domain, where low-spectrum components capture coarse structural information, while high-spectrum components encode fine-grained geometric details:
\begin{equation}
   \hat{S'} = \phi_{GFT}(S') = U^T S',
\label{eq:GFT}
\end{equation}
where $\hat{S'} \in \mathbb{R}^{k_g\times3}$. 
Correspondingly, the Inverse GFT (IGFT) reconstructs the spatial patch from the spectral domain: 
\begin{equation}
    S' = \phi_{IGFT}(\hat{S}) = U \hat{S'}.
\label{eq:IGFT}
\end{equation}
An optimizable spectral trigger $\xi \in \mathbb{R}^{k_g\times3}$ is applied to each selected patch in the spectral domain to generate the poisoned patches $S^p$:
\begin{equation}
  S^p=\phi_{IGFT}(\phi_{GFT}(S')+\xi).
  \label{eq:attack}
\end{equation}
The final poisoned point cloud sample $X^p$ is then constructed by replacing the original selected patches with the reconstructed patches while keeping the unselected regions unchanged:
\begin{equation}
   X^p=\mathcal{T}(X)=(X \setminus \{S'_1, S'_2, ..., S'_m\}) \cup \{S_1^p, S_2^p, ..., S_m^p\}.
\end{equation}
Since KNN neighborhoods may overlap, a point can appear in multiple selected patches.
SPBA uses a deterministic first-assignment rule in the selected-patch order: if a point has already been updated by an earlier selected patch, later reconstructed patches do not overwrite this assignment.

\begin{figure}[htbp]
    \centering
    \includegraphics[width=1.0\linewidth]{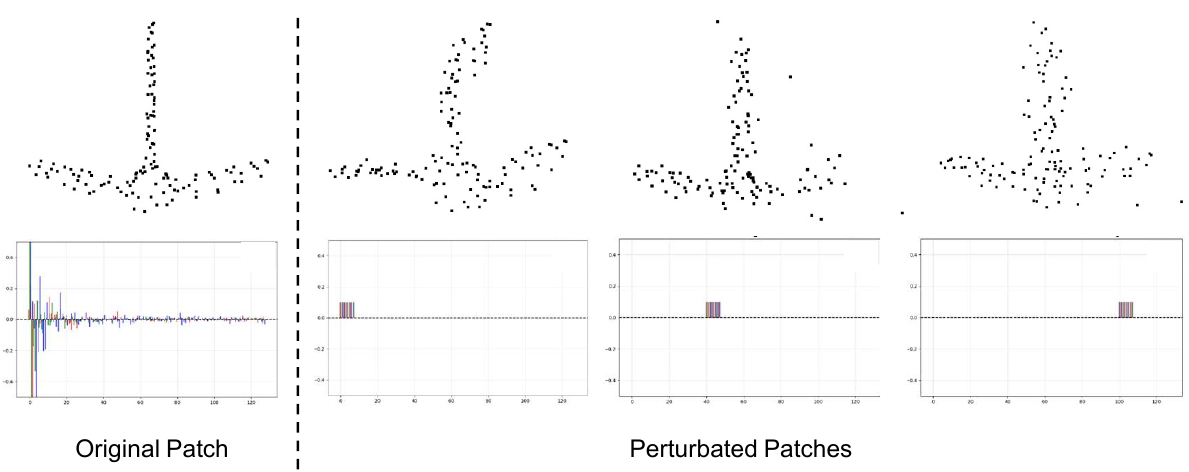}
    \caption{Comparison of original and perturbed patches of airplane tail. The second row shows the spectral features of the original patch and spectral perturbations across different spectrum bands.}
    \label{fig:patch_freq}
\end{figure}

\subsection{Optimization Strategy}

Fig.~\ref{fig:patch_freq} illustrates the frequency-domain trade-off behind SPBA. Strong low-frequency perturbations distort the overall patch geometry, whereas overly strong high-frequency perturbations tend to create isolated artifacts. The optimization therefore aims to learn a trigger that activates the backdoor while keeping the reconstructed patch close to the original geometry.
The backdoor learning process follows an alternating optimization scheme, where the model parameters $\theta$  and the spectral trigger $\xi$ are updated iteratively, with one being optimized while the other remains frozen. Thus, each optimization step consists of two forward and backward propagations.

To train the model parameters $\theta$, we define the loss function $L_M$, following the standard formulation for backdoor attacks:
\begin{equation}
L_{M} =\sum_{D_b} L_{CE} (f_{\theta}(X), Y) + \sum_{D_p}L_{CE}(f_{\theta}(X^p), Y_t),
\end{equation}
where $L_{CE}$ is the cross-entropy loss. The first term preserves classification performance on benign data, and the second term fits the poisoned samples to the target label.

To optimize the patch-wise trigger $\xi$ for both attack efficiency and high stealthiness, we ensure that poisoned samples are misclassified into the target class $Y_t$ while maintaining minimal perceptual distortions. This is achieved through the following loss function $L_T$:
\begin{equation}
L_{T} = \sum_{ D} \left( L_{CE}(f_{\theta}(\mathcal{T}(X)), Y_t) +  L_{reg}(X, \mathcal{T}(X)) \right).
\end{equation}
The cross-entropy term encourages the trigger to induce the target prediction reliably. The regularizer $L_{reg}$ keeps the poisoned geometry close to the original geometry.
In the regularizer function $L_{reg}$, we employ Euclidean distance $L_{ED}(X, \mathcal{T}(X))$ and Chamfer distance  $L_{CD}(X, \mathcal{T}(X))$ to enforce similarity between the original and poisoned point clouds, and Hausdorff distance $ L_{HD}(X, \mathcal{T}(X))$ is used to control the worst scenario of outliers.
The final regularization term is formulated as:
\begin{equation}
    L_{reg} = \lambda_1 L_{ED}+ \lambda_2  L_{CD}+ \lambda_3 L_{HD},
\label{eq:regularizer}
\end{equation}
where $\lambda_1,\lambda_2,\lambda_3$ control the relative contribution of each loss term.
This joint optimization framework ensures that the spectral trigger remains both effective and imperceptible, enabling stealthy backdoor attacks on 3D point cloud models.

\section{Experiments and Results}
\subsection{Experiment Details}
\textbf{Datasets and Models.}

The evaluation uses two widely adopted 3D point cloud classification benchmarks. ModelNet40~\cite{ModelNet} contains 40 object categories with 9,843 training samples and 2,468 testing samples. ShapeNetPart~\cite{ShapeNet} contains 12,128 training samples and 2,874 testing samples across 16 object categories. For both datasets, 1,024 points are uniformly sampled from each shape together with their normal vectors, and the coordinates are normalized for consistency. Following prior 3D backdoor work~\cite{li2021pointba,gao2023IRBA}, the evaluation covers four representative architectures: PointNet~\cite{qi2017pointnet}, PointNet++~\cite{qi2017pointnet++}, DGCNN~\cite{wang2019DGCNN}, and PointNext~\cite{qian2022pointnext}.

\noindent\textbf{Attacks Setup.}
We compare SPBA against four representative point cloud attack methods: PointBA-I~\cite{li2021pointba}, PointBA-O~\cite{li2021pointba}, IRBA~\cite{gao2023IRBA}, and IBAPC~\cite{fan2024IBAPC}. 
PointBA-I is reproduced by implanting a spherical cluster that contains 3\% of the total points, with radius 0.05 and center $(0.5,0.5,0.5)$ in the coordinate space. PointBA-O applies a $10^{\circ}$ rotation along the Z-axis. The default configurations from the original papers are used for IRBA and IBAPC. For SPBA, $g$ and $k_g$ are both set to 32 for patch decomposition, and $m=16$ patches are selected for trigger injection. The regularizer in Eq.~\eqref{eq:regularizer} uses $\lambda_1=1$, $\lambda_2=5$, and $\lambda_3=1$. For PointNet, $\lambda_1$ is set to 0.1 to improve training stability.

For all attack methods, we set the poisoning rate $\rho=0.1$. The target class is chosen as Chair ($Y_t=8$) for ModelNet40 and Lamp ($Y_t=8$) for ShapeNetPart. 
We employ two Adam optimizers for backdoored model training and spectral trigger optimization, with learning rates of 0.001 and 0.01, respectively, and a weight decay of 1e-4. A cosine annealing learning rate scheduler is applied to both optimizers to facilitate stable convergence. Training is performed for 100 epochs with a batch size of 32. All experiments are conducted on a workstation equipped with an NVIDIA RTX 4090 GPU.

\noindent\textbf{Evaluation Metrics.}

The evaluation reports Benign Accuracy (BA), Attack Success Rate (ASR), and Chamfer Distance (CD). BA measures the classification accuracy on benign test samples. ASR measures the fraction of poisoned test samples classified into the target label. CD quantifies the geometric distortion between poisoned and original point clouds. A desirable attack should achieve high ASR with low CD while preserving BA as much as possible. For clarity, both BA and ASR are presented as percentages, and CD values in tables are \textbf{scaled by a factor of 1,000}.

\setlength{\tabcolsep}{2.5pt}
\begin{table*}[htb]
  \centering
  \caption{Comparison of attack performance across different models on ModelNet40. BA and ASR are presented as percentages, and CD values in tables are scaled by a factor of 1,000. The best results are \textbf{bolded} and the second-best results are \underline{underlined}.}
  \resizebox{1.0\textwidth}{!}{
    \begin{tabular}{ccccccccccccccccc}
    \toprule
    
     \multirow{2}[2]{*}{Model} & Clean & \multicolumn{3}{c}{PointBA-I} & \multicolumn{3}{c}{PointBA-O} & \multicolumn{3}{c}{IRBA} & \multicolumn{3}{c}{IBAPC} & \multicolumn{3}{c}{SPBA} \\
     \cmidrule(lr){3-5} \cmidrule(lr){6-8} \cmidrule(lr){9-11} \cmidrule(lr){12-14} \cmidrule(lr){15-17}         
          & BA$\uparrow$    & BA$\uparrow$    & ASR$\uparrow$   & CD$\downarrow$    & BA$\uparrow$    & ASR$\uparrow$   & CD$\downarrow$    & BA$\uparrow$    & ASR$\uparrow$   & CD$\downarrow$    & BA$\uparrow$    & ASR$\uparrow$   & CD$\downarrow$    & BA$\uparrow$    & ASR$\uparrow$   & CD$\downarrow$ \\
    \midrule
    \midrule
    PointNet & 89.14  & \textbf{89.67} & \textbf{99.03} & 4.08  & 88.17  & 94.77  & 5.31  & 88.25  & 93.27  & 18.54  & 88.21  &  91.57  & {\underline{3.04}} & \underline{88.90} & \underline{97.89} & \textbf{2.60} \\
    PointNet++ & 90.80  & \underline{90.36} & \textbf{98.87} & 4.08  & 89.55  & 96.68  & 5.31  & \textbf{90.40} & \underline{98.26} & 18.54  & 89.67  & 95.58  & \underline{0.68} & 89.42  & 97.97  & \textbf{0.11} \\
    DGCNN & 89.75  & \textbf{91.05} & \textbf{100.00} & 4.08  & 89.67  & 95.02  & 5.31  & 89.55  & 92.38  & 18.54  & 88.24  & 95.44  & \underline{0.87} & \underline{89.75} & \underline{97.61} & \textbf{0.69} \\
    PointNext & 89.22  & 88.21  & 95.83  & 4.08  & \textbf{88.94} & 96.31  & 5.31  & 88.41  & 92.46  & 18.54  & \underline{88.70} & \underline{96.60} & \underline{1.48} & 88.62  & \textbf{96.85} & \textbf{0.92} \\
    \midrule
    \midrule
    Average   & 89.73  & \textbf{89.82} & \textbf{98.43} & 4.08  & 89.08  & 95.69  & 5.31  & 89.15  & 94.09  & 18.54  & 88.70  & 94.80  & \underline{1.52} & \underline{89.17} & \underline{97.58} & \textbf{1.08} \\
    \bottomrule
    \end{tabular}%
    }

  \label{tab:modelnet40}%
\end{table*}%

\subsection{Attack Effectiveness}
The results presented in Table~\ref{tab:modelnet40} provide a comparative evaluation of various backdoor attack methods on ModelNet40. 
As shown in Table~\ref{tab:modelnet40}, our proposed SPBA achieves the lowest average CD value of 1.08 among all attacks, demonstrating its state-of-the-art stealthiness. Moreover, SPBA outperforms most backdoor attacks according to BA and ASR.  In particular, compared to the sample-wise spectral attack IBAPC, SPBA achieves notable improvements of 0.47\% in average BA and 2.78\% in average ASR, highlighting the effectiveness of our patch-wise spectral trigger.
Although PointBA-I achieves the highest average ASR of 98.43\%, it comes at the cost of a significantly higher average CD of 4.08, making it much more detectable.  Subsequent defense experiments further confirm that the trigger used in PointBA-I is especially susceptible to detection or removal by existing defense methods. 
Table~\ref{tab:shapenet} shows more experimental results on the ShapeNetPart dataset. 
SPBA achieves a similar performance, offering the best balance between attack effectiveness and stealthiness. These results further validate the superiority of SPBA as a more effective and imperceptible backdoor attack method compared to existing attacks.

\setlength{\tabcolsep}{2.5pt}
\begin{table*}[htb]
  \centering
      \caption{Comparison of attack performance across different models on ShapeNetPart. BA and ASR are presented as percentages, and CD values in tables are scaled by a factor of 1,000. The best results are \textbf{bolded} and the second-best results are \underline{underlined}.}
    \resizebox{1.0\textwidth}{!}{
    \begin{tabular}{ccccccccccccccccc}
    \toprule
    \multirow{2}[2]{*}{Model} & Clean & \multicolumn{3}{c}{PointBA-I} & \multicolumn{3}{c}{PointBA-O} & \multicolumn{3}{c}{IRBA} & \multicolumn{3}{c}{IBAPC} & \multicolumn{3}{c}{SPBA} \\   
    \cmidrule(lr){3-5} \cmidrule(lr){6-8} \cmidrule(lr){9-11} \cmidrule(lr){12-14} \cmidrule(lr){15-17}  
          & BA$\uparrow$ & BA$\uparrow$    & ASR$\uparrow$   & CD$\downarrow$    & BA$\uparrow$    & ASR$\uparrow$   & CD$\downarrow$    & BA$\uparrow$    & ASR$\uparrow$   & CD$\downarrow$    & BA$\uparrow$    & ASR$\uparrow$   & CD$\downarrow$    & BA$\uparrow$    & ASR$\uparrow$   & CD$\downarrow$ \\
    \midrule
    \midrule
    PointNet & 98.47  & \textbf{98.30} & \textbf{99.72} & 7.48  & 97.95  & 98.34  & 5.67  & \underline{98.12} & 96.80  & 16.06  & 97.80  & 96.01  & \underline{1.73} & \textbf{98.30} & \underline{98.40} & \textbf{1.22} \\
    PointNet++ & 98.99  & 98.75  & \textbf{99.69} & 7.48  & \underline{98.50} & 98.78  & 5.67  & \textbf{98.89} & \underline{99.03} & 16.06  & 98.43  & 98.46  & \underline{0.22} & 98.26  & 98.30  & \textbf{0.02} \\
    DGCNN & 99.06  & \textbf{98.92} & \textbf{100.00} & 7.48  & \underline{98.85} & 98.50  & 5.67  & 98.68  & 97.63  & 16.06  & 97.54  & 95.44  & \underline{0.80} & 98.43  & \underline{99.23} & \textbf{0.44} \\
    PointNext & 98.82  & \textbf{98.61} & \textbf{99.51} & 7.48  & 98.10 & 96.83  & 5.67  & 96.31 & 96.17  & 16.06  & 97.35  & 95.99  & \underline{0.44} & \underline{98.53}  & \underline{98.45} & \textbf{0.37} \\
    \midrule
    \midrule
    Average   & 98.83  & \textbf{98.64} & \textbf{99.73} & 7.48  & 98.35 & 98.11  & 5.67  & 98.00  & 97.41  & 16.06  & 97.78  & 96.48  & \underline{0.80} & \underline{98.38}  & \underline{98.60} & \textbf{0.51} \\
    \bottomrule
    \end{tabular}%
    }

  \label{tab:shapenet}%
\end{table*}%

\subsection{Attack Stealthiness}
\begin{figure*}[htb]
    \centering
    \includegraphics[width=0.95\linewidth]{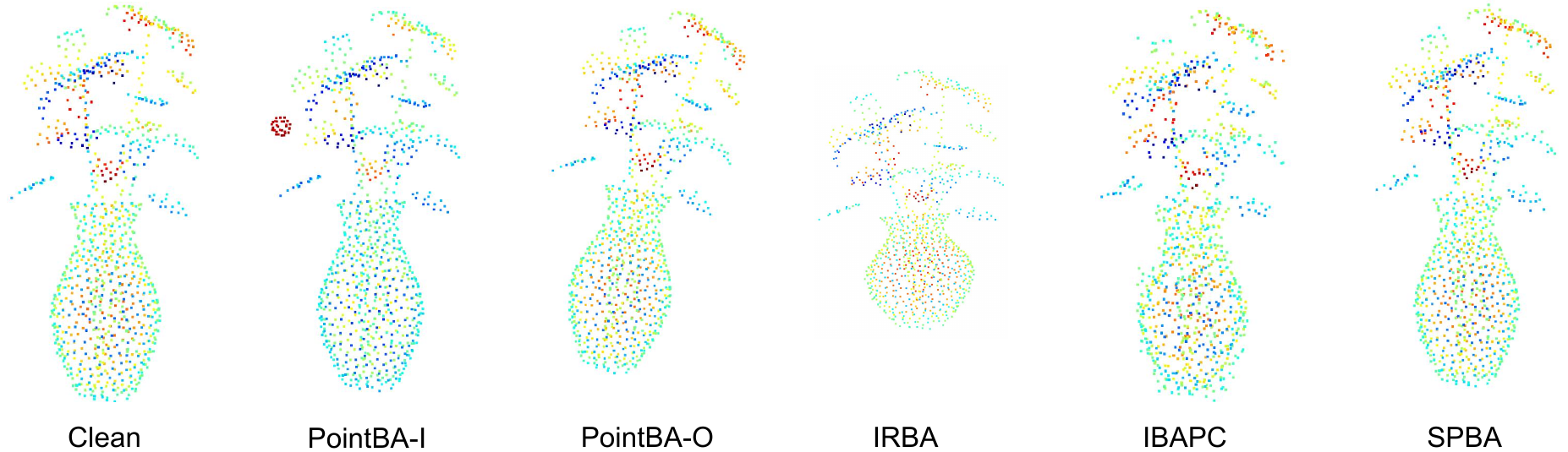}
    \caption{Visual comparison of poisoned samples from different backdoor attack methods. Our proposed SPBA preserves structural integrity and ensures smooth regions (e.g., the vase or car roof) remain undisturbed, enhancing imperceptibility. }
    \label{fig:visualization}
\end{figure*}

We visualize some poisoned samples generated by different backdoor attack methods in  Fig.~\ref{fig:visualization}. As observed,  PointBA-I introduces noticeable anomalies compared to clean samples, significantly increasing its detectability.
Although PointBA-O and IRBA generate natural-looking poisoned samples by applying transformation, they still cause considerable deformations in shape.
Compared to IBAPC, SPBA selectively perturbs the flower region while preserving the smooth structure of the vase, making the modifications less perceptible to human observers.

Furthermore, we conduct a human perceptual study to assess the stealthiness of the proposed attack. Following the protocol in IRBA~\cite{gao2023IRBA}, we present a mixture of clean and poisoned samples to 20 participants, all of whom are over 18 years old, major in computer science, and have prior knowledge of machine learning and computer vision.
Each participant is shown 10 groups of multi-view renderings of point clouds from distinct object classes. Each group consists of one clean reference sample together with five attack types of its poisoned version arranged in random order. Participants are instructed to assign a score $\in \{1, 2, 3, 4, 5\}$ to indicate the likelihood that the poisoned samples in the group have been modified. A lower score reflects a lower perceived likelihood, suggesting that the poisoned samples are less perceptible.
The average score for each attack method is reported in Table~\ref{tab:human}. The results demonstrate that SPBA achieves the highest level of stealthiness, yielding the lowest perceptibility score among all compared methods.

\begin{table}[htbp]

  \centering
  \caption{Scores of the human perceptual study.}
    \begin{tabular}{cccccc}
    \toprule
    Method & PointBA-I & PointBA-O & IRBA  & IBAPC & SPBA \\
    \midrule
    \midrule
    Score $\downarrow$ &   4.79    &  3.33     & 4.05      &   2.76    & \textbf{1.53} \\
    \bottomrule
    \end{tabular}%
    
  \label{tab:human}%
  
\end{table}%

\subsection{Attack Efficiency}
Eq.~\eqref{eq:attack} shows that the computational cost of spectral attacks is highly dependent on the trigger size. We further report the computational cost of different spectral attacks in Table~\ref{tab:computational_cost}. The MFLOPs are obtained by profiling the complete trigger-injection pipeline implemented in their official code during inference with a batch size of 64, rather than by estimating only the nominal trigger tensor size.
The prior spectral attack IBAPC uses a large trigger of size $\xi^{\text{IBAPC}} \in \mathbb{R}^{1024 \times 3}$, resulting in a substantial computational cost of 939.72 MFLOPs, which severely hinders its efficiency and practicality for deployment. 
In contrast, although our proposed SPBA perturbs an average of 452 points, which accounts for 44\% of the sample, our patch-based spectral trigger is much smaller ($\xi^{SPBA} \in \mathbb{R}^{32 \times 3}$). The measured computational cost of SPBA is only 14.74 MFLOPs, which is comparable to that of a single fully connected layer.

For a fairer same-scale reference, an FPS-based sample-wise spectral trigger (FPSP, $\in \mathbb{R}^{452 \times 3}$) is also constructed to perturb approximately the same number of points as SPBA. Even under this matched perturbation budget, FPSP is 11.54$\times$ more expensive than SPBA and performs worse in Table~\ref{tab:select2}. These results indicate that the efficiency gain comes from the local trigger design rather than simply from perturbing fewer points.

\setlength{\tabcolsep}{12pt} 
\begin{table}[htbp]

  \centering
  
    \caption{Computational cost of spectral trigger injection.}
    \begin{tabular}{cccc}
    \toprule
    Method & IBAPC &  FPSP   & SPBA \\
    \midrule
    \midrule
    MFLOPs $\downarrow$ & 939.72 & 170.06 & \textbf{14.74} \\
    \bottomrule
    \end{tabular}%
    
  \label{tab:computational_cost}%
\end{table}%

\subsection{Resistance to Defense Methods}

\noindent\textbf{Resistance to Data Augmentation.}
Since data augmentation is widely employed to enhance the resistance of classification models, we introduce six types of augmentations to evaluate their impact on backdoor attack performance:
1) Rotation: random rotation around the z-axis with a maximum angle of $10^{\circ}$; 2) Rotation3D: random rotation up to $10^{\circ}$ along all three axes; 
3) Scaling: scaling within a range of 0.5 to 1.5; 
4) Shift: shifting randomly between -0.1 and 0.1 along each axis; 
5) Dropout:  removing up to 20\% of points; and 
6) Jitter: introducing noise sampled from $N(0,0.02)$. 
The experiments are conducted on the ModelNet40 dataset using the PointNet++ model.
As shown in Table~\ref{tab:augmentation}, different augmentations affect attack methods to varying degrees.
Due to the high visibility of the trigger of PointBA-I, augmentations have minimal impact, as the trigger remains distinct. 
However, for rotation-based PointBA-O, applying Rotation significantly reduces attack effectiveness.
For IRBA, which relies on local transformations, augmentations such as Rotation3D, Dropout and Jitter cause a notable decline in attack performance, with ASR dropping by up to 8.23\%.
Conversely, for optimization-based attacks like IBAPC and our proposed SPBA, only Jitter leads to a slight ASR reduction, while other augmentations have negligible effects, demonstrating their resistance against most augmentations.

\setlength{\tabcolsep}{2pt} 
\begin{table*}[htbp]

  \centering
  \caption{Resistance of backdoor attacks to varied data augmentations and SOR.}
    \begin{tabular}{ccccccccccc}
    \toprule
    \multirow{2}[2]{*}{Augmentation} & \multicolumn{2}{c}{PointBA-I} & \multicolumn{2}{c}{PointBA-O} & \multicolumn{2}{c}{IRBA} & \multicolumn{2}{c}{IBAPC} & \multicolumn{2}{c}{SPBA} \\
    \cmidrule(lr){2-3} \cmidrule(lr){4-5} \cmidrule(lr){6-7} \cmidrule(lr){8-9} \cmidrule(lr){10-11} 
          & BA$\uparrow$   & ASR$\uparrow$   & BA$\uparrow$   & ASR$\uparrow$   & BA$\uparrow$  & ASR$\uparrow$   & BA$\uparrow$   & ASR$\uparrow$   & BA$\uparrow$  & ASR$\uparrow$ \\
    \midrule
    \midrule
    No    & 90.36  & 98.87  & 89.55  & 96.68  & 90.40  & 98.26  & 89.67  & 95.58  & 89.42  & 97.97  \\
    Rotation     & 89.42  & 98.95  & 86.83  & 7.66  & 90.52  & 97.69  & 90.59  & 95.90  & 89.34  & 98.22  \\
    Rotation3D    & 89.71  & 99.11  & 88.78  & 30.51  & 90.68  & 92.79  & 90.23  & 95.94  & 89.38  & 97.24  \\
    Scaling & 90.72  & 99.55  & 91.65  & 96.92  & 91.17  & 99.47  & 90.85  & 96.47  & 91.57  & 98.26  \\
    Shift & 90.11  & 99.03  & 89.67  & 96.07  & 90.03  & 97.93  & 89.74  & 94.23  & 89.59  & 97.08  \\
    Dropout & 90.52  & 99.39  & 90.07  & 96.11  & 90.40  & 94.85  & 88.37  & 94.93  & 90.76  & 97.24  \\
    Jitter & 89.42  & 98.70  & 89.34  & 95.06  & 90.07  & 90.03  & 89.91  & 91.65  & 90.96  & 95.01  \\
    \midrule
    SOR & 90.36 & 46.47 & 89.55& 95.46 & 90.40 & 97.30 & 89.67 & 95.13 & 89.42 & 97.36\\
    \bottomrule
    \end{tabular}%
    
  \label{tab:augmentation}%
\end{table*}%

\noindent\textbf{Resistance to SOR.}
Statistical Outlier Removal (SOR)~\cite{zhou2019dup} is designed to eliminate outliers based on the local neighborhood density using the KNN algorithm. In our implementation, we set KNN to use 60 nearest neighbors and set the standard deviation threshold to 0.5.
We apply it during inference on PointNet++ models trained on ModelNet40. 
The results in Table~\ref{tab:augmentation} show that only PointBA-I experiences a significant drop in ASR, decreasing from 98.87\% to 46.47\%, while the ASR reduction for other attacks remains within 2\%.
This suggests that SOR serves as an effective defense specifically against injection-based attacks while having a limited impact on other attack methods.

\begin{figure}[tbp]
    \centering
    \includegraphics[width=0.8\linewidth]{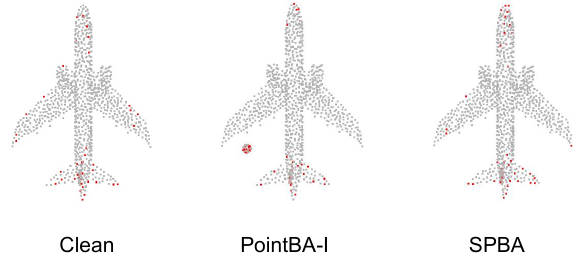}
    \caption{Gradient-based salience analysis with the most significant points highlighted in red.}
    \label{fig:saliency}
\end{figure}

\noindent\textbf{Resistance to the Gradient-based Method.}
As many studies on backdoor attacks assess trigger stealthiness using gradient-based methods~\cite{selvaraju2017gradcam}, we employ a point cloud saliency map~\cite{zheng2019saliencymap} to identify salient points that contribute most to the model decision process. Specifically, 40 points with the highest saliency scores are marked in red. As illustrated in Fig.~\ref{fig:saliency}, the spherical trigger of PointBA-I is easily detected. In contrast, our proposed SPBA exhibits a saliency distribution similar to the clean model, indicating enhanced stealthiness and reduced detectability.

\noindent\textbf{Resistance to PointCRT.}
PointCRT~\cite{hu2023pointcrt} utilizes a nonlinear classifier to detect the abnormal corruption resistance of poisoned samples. The F1 score (F1) and Area Under the Curve (AUC) are employed to evaluate the ability of PointCRT to detect the attack,  where lower values indicate that the attack is more resistant to detection by PointCRT.
As shown in Table~\ref{tab:PointCRT}, SPBA outperforms almost all the attacks except PointBA-O, which exhibits limited attack performance and is vulnerable to data augmentations. 

\setlength{\tabcolsep}{3pt} 
\begin{table}[htbp]
  \centering
  \caption{Performance of PointCRT on different attacks.}
    \begin{tabular}{cccccc}
    \toprule
    Method & PointBA-I & PointBA-O & IRBA  & IBAPC & SPBA \\
    \midrule
    \midrule
    F1 $\downarrow$   & 95.94 & \textbf{72.89} & 93.98 & 92.84 & \underline{87.76} \\
    AUC $\downarrow$  & 97.52 & \textbf{77.76} & 95.75 & 92.93 & \underline{87.89} \\
    \bottomrule
    \end{tabular}%
    
  \label{tab:PointCRT}%
\end{table}%

\noindent\textbf{Resistance to FLARE.}
FLARE~\cite{hou2025flare} aggregates abnormal activations across hidden layers and adaptively selects the most stable subspace to divide samples into two clusters, detecting the more stable cluster as poisoned samples. Since this method is independent of input modality, it could also be applied to point cloud backdoor sample detection. Specifically, we evaluated the method on backdoored DGCNN models trained on ModelNet40. The true positive rate (TPR) reflects the proportion of correctly identified poisoned training samples, with a lower TPR indicating weaker detection capability. Experimental results in Table~\ref{tab:FLARE} show that our approach consistently yields the lowest TPR, demonstrating strong resistance against FLARE.

\setlength{\tabcolsep}{3pt}
\begin{table}[htbp]
  \centering
  \caption{Detection accuracy of FLARE on different attacks.}
    \begin{tabular}{cccccc}
    \toprule
    Method & PointBA-I & PointBA-O & IRBA  & IBAPC & SPBA \\
    \midrule
    \midrule
    TPR $\downarrow$   & 56.93 & 62.01 & 47.97 & 38.01 & \textbf{29.78} \\
    \bottomrule
    \end{tabular}%
    
  \label{tab:FLARE}%
\end{table}%

\subsection{Ablation Study}

\noindent\textbf{Effect of Patch Selection Strategies.}
In our default setting, we select the 16 patches with the highest patch imperceptibility scores (PIS), denoted as H-PIS. To evaluate the effectiveness of this selection strategy, we compare H-PIS with three alternative patch selection strategies:
(1) L-PIS: Select the patches with the lowest PIS;
(2) Random: Randomly select the patches;
(3) FPSP: Use the FPS algorithm to sample the same number of points as H-PIS.
We evaluate their average performance on ModelNet40. As shown in Table~\ref{tab:select2},  the H-PIS strategy achieves the highest attack effectiveness and the strongest stealthiness. This result underscores the advantage of selecting high-imperceptibility patches for local spectral trigger injection.
\begin{table}[htbp]

  \centering
  \caption{ Comparison of attack performance of SPBA with various selection strategies.}
    \begin{tabular}{ccccc}
    \toprule
    Method & L-PIS & Random & FPSP   & H-PIS (default) \\
    \midrule
    \midrule
    BA $\uparrow$   & 87.99 & 88.19 & 87.84 & \textbf{89.17} \\
    ASR $\uparrow$   & 95.53 & 96.35 & 96.45 & \textbf{97.58} \\
    CD $\downarrow$   & 1.16  & 1.29  & 1.39  & \textbf{1.08} \\
    \bottomrule
    \end{tabular}%
    
  \label{tab:select2}%
\end{table}%

\noindent\textbf{Effect of Selected Patch Number $m$.}
In this section, we experiment with injecting the spectral trigger into varying numbers $m$ of selected patches during the training process.
The experiments are conducted on ModelNet40 using PointNet++.
As shown in Table~\ref{tab:patchnum}, as $m$ increases from 1 to 16, the CD value gradually decreases, indicating improved stealthiness. This suggests that selecting more patches within a reasonable range helps distribute perturbations more naturally, enhancing imperceptibility. However, when $m$ further increases to 32, where almost all points are perturbed, CD rises to 0.12, indicating that excessive modifications start to introduce noticeable distortions, reducing stealthiness. Based on these observations, we set $m = 16$ as the default configuration, as it achieves the optimal balance between imperceptibility (lowest CD), strong attack effectiveness, and benign accuracy.
\begin{table}[htbp]
  \centering
  \caption{The effect of selected patch number $m$.}
    \begin{tabular}{cccccccc}
    \toprule
    $m$     & 1     & 2     & 4     & 8     & 16  & 24  & 32 \\
    \midrule
    \midrule
    BA    & 89.56  & 89.88  & 88.53  & \textbf{90.15} & 89.42  & 89.27 & 89.36  \\
    ASR   & 97.38  & 97.10  & 97.28  & 96.84  & \textbf{97.97} & 97.73 & 97.30  \\
    CD    & 0.17  & 0.19  & 0.18  & 0.14  & \textbf{0.11} & \textbf{0.11}  & 0.12 \\
    \bottomrule
    \end{tabular}%
    
  \label{tab:patchnum}%
\end{table}%

\noindent\textbf{Effect of the Patch Size.}
We conduct additional experiments on ModelNet40 using PointNet++ to further investigate the effect of patch size. To ensure that the total number of perturbed points remains approximately constant, when the patch size is doubled, the number of selected patches is reduced to half, and vice versa. From Table~\ref{tab:patch_size}, ASR attains its maximum at $k_g=32$ and CD is minimized at this setting, offering the best balance of attack success and stealth. The benign accuracy is still high (89.42\%), ranking second among the evaluated patch sizes.


\begin{table}[htbp]
  \centering
    \caption{The effect of patch size $k_g$}
    \begin{tabular}{ccccc}
    \toprule
    $k_g$  & 16    & 32    & 64    & 128 \\
    \midrule
    \midrule
    BA    & 89.30  & 89.42  & \textbf{89.75}  & 89.26 \\
    ASR   & 97.00  & \textbf{97.97}  & 97.64  & 97.73 \\
    CD    & 0.13  & \textbf{0.11}  & 0.18  & 0.15 \\
    \bottomrule
    \end{tabular}%
    
  \label{tab:patch_size}%
\end{table}%

\begin{figure}[htbp]
  \centering
  \begin{subfigure}{0.5\textwidth}
    \centering
    \includegraphics[width=\linewidth]{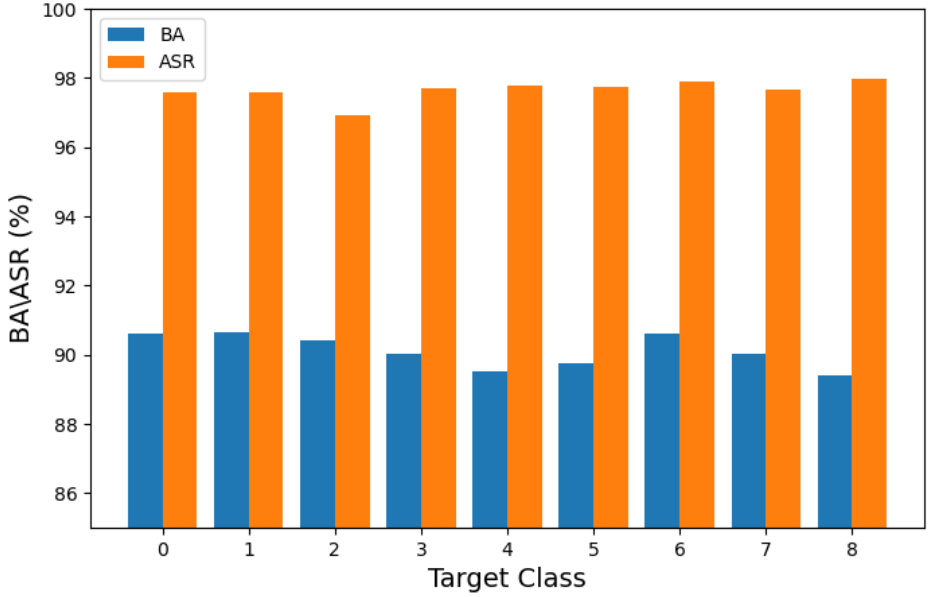} 
    \caption{The effect of the target class selection.}
    \label{fig:sub1}
  \end{subfigure}\hfill
  \begin{subfigure}{0.5\textwidth}
    \centering
    \includegraphics[width=\linewidth]{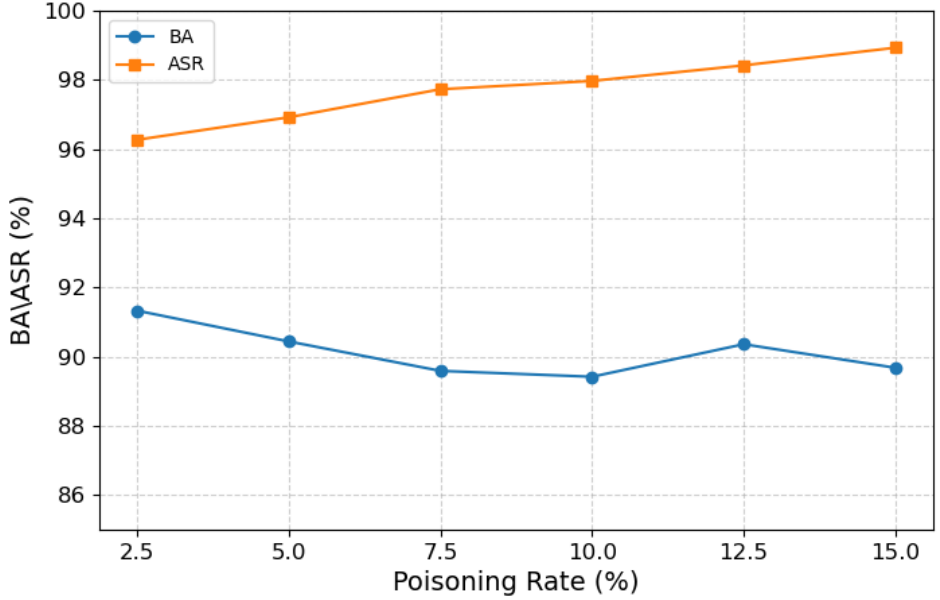}
    \caption{The effect of poisoning rate.}
    \label{fig:sub2}
  \end{subfigure}

  \caption{The effect of the target class selection and poisoning rate.}
  \label{fig:label}
\end{figure}

\noindent\textbf{Effect of the Target Class.}
In this section, we investigate the effect of different target classes on our attack performance. Specifically, we select the top 8 categories from ModelNet40 as target classes and conduct experiments using PointNet++.
As shown in Fig.~\ref{fig:label}(a), the choice of target class has a minimal effect on both Attack Success Rate (ASR) and Benign Accuracy (BA), with fluctuations not exceeding 1\%. These results demonstrate the robustness of our approach across various target label settings, ensuring its consistency in different attack scenarios.

\noindent\textbf{Effect of the Poisoning Rate.}
We explore the impact of the poisoning rate on our attack by conducting experiments on ModelNet40 using PointNet++. As shown in Fig.~\ref{fig:label}(b), ASR consistently increases as the poisoning ratio rises. BA exhibits minor fluctuations but remains relatively stable as the poisoning rate increases.  Notably, even at a low poisoning rate of 2.5\%, our approach achieves an ASR of 96.27\%, underscoring its effectiveness even with limited poisoned data.

\noindent\textbf{Effect of the Regularization Loss Weights.}
In this section, we investigate the effect of the regularization loss weights in Eq.~\ref{eq:regularizer}, where $\lambda_1$, $\lambda_2$, and $\lambda_3$ control the contributions of $L_{ED}$, $L_{CD}$, and $L_{HD}$, respectively. As shown in Table~\ref{tab:loss}, increasing $\lambda_1$ gradually enhances the stealthiness (reflected by lower CD values), but at the cost of a decreasing ASR. Notably, when $\lambda_1=1$, the attack achieves a relatively good trade-off between aggressiveness and stealthiness. Moreover, our proposed SPBA algorithm demonstrates robustness with respect to the hyperparameters $\lambda_2$ and $\lambda_3$. Based on these observations, we ultimately set $\lambda_2=5$ and $\lambda_3=1$ to achieve a favorable balance between attack effectiveness and stealthiness.
\begin{table}[htbp]
  \centering
  \caption{The effect of the regularization loss weights}
    \begin{tabular}{cccccc}
    \toprule
    $\lambda_1$ & $\lambda_2$ & $\lambda_3$ & BA    & ASR   & CD \\
    \midrule
    \midrule
    1     & 5     & 1     & 89.42 & 97.97 & 0.11 \\
    0.1   & 5     & 1     & 89.72 & 99.39 & 0.72 \\
    0.5   & 5     & 1     & 89.47 & 98.50  & 0.19 \\
    2     & 5     & 1     & 87.64 & 92.70  & 0.08 \\
    1     & 1     & 1     & 89.95 & 97.17 & 0.15 \\
    1     & 3     & 1     & 89.34 & 97.77 & 0.11 \\
    1     & 5     & 0.1   & 89.74 & 96.96 & 0.12 \\
    1     & 5     & 0.5   & 89.62 & 97.44 & 0.11 \\
    1     & 5     & 2     & 89.52 & 97.71 & 0.11 \\
    \bottomrule
    \end{tabular}%
    
  \label{tab:loss}%
\end{table}%

\section{Discussion and Limitations}
SPBA points to a useful design direction for 3D backdoor attacks. A trigger need not be global to remain transferable, nor sample-specific to exploit local geometry effectively. By shifting trigger optimization from the full point set to selected local spectral neighborhoods, the attack becomes easier to regularize and more efficient to optimize. The ablation results further suggest that the observed trade-off is driven by the combination of geometry-aware patch selection and spectral design, rather than by either component in isolation.

Several limitations should nevertheless be noted. First, the current evaluation is restricted to normalized shape classification benchmarks, namely ModelNet40 and ShapeNetPart, which do not fully capture the density variation and sensor noise present in real scanned point clouds. Second, the trigger is purely digital, and physical realizability under real sensor acquisition remains beyond the scope of this work. Addressing these limitations through more realistic scanned-data evaluation and physically grounded attack settings represents an important direction for future research.

\section{Conclusion}

In this paper, we proposed \emph{Stealthy Patch-Wise Backdoor Attack} (SPBA), a patch-wise spectral backdoor attack for 3D point clouds. By defining trigger support on geometrically constructed local patches and optimizing the trigger in the spectral domain, SPBA enables localized perturbation injection while better preserving the overall shape and original input format. A curvature-based Patch Imperceptibility Score (PIS) was further introduced to identify geometrically complex regions for stealthier trigger placement.
Experiments on ModelNet40 and ShapeNetPart showed that, under the current benchmark setting, SPBA achieved the best stealthiness among the compared methods while maintaining competitive attack performance and significantly reducing the computational cost of spectral-trigger optimization.

{
    \small
    \bibliographystyle{ieeenat_fullname}
    \bibliography{main}

@String(PAMI = {IEEE Trans. Pattern Anal. Mach. Intell.})

@String(CVPR= {IEEE Conf. Comput. Vis. Pattern Recog.})

@String(ICCV= {Int. Conf. Comput. Vis.})

@String(ECCV= {Eur. Conf. Comput. Vis.})

@String(TOG= {ACM Trans. Graph.})

@String(ICASSP=	{ICASSP})

@String(AAAI = {AAAI})

@String(PAMI  = {IEEE TPAMI})

@String(CVPR  = {CVPR})

@String(ICCV  = {ICCV})

@String(ECCV  = {ECCV})

@String(TOG   = {ACM TOG})

@inproceedings{ModelNet,
  title={3d shapenets: A deep representation for volumetric shapes},
  author={Wu, Zhirong and Song, Shuran and Khosla, Aditya and Yu, Fisher and Zhang, Linguang and Tang, Xiaoou and Xiao, Jianxiong},
  booktitle={Proceedings of the IEEE/CVF Conference on Computer Vision and Pattern Recognition (CVPR)},
  pages={1912--1920},
  year={2015}
}

@article{ShapeNet,
  title={A scalable active framework for region annotation in 3d shape collections},
  author={Yi, Li and Kim, Vladimir G and Ceylan, Duygu and Shen, I-Chao and Yan, Mengyan and Su, Hao and Lu, Cewu and Huang, Qixing and Sheffer, Alla and Guibas, Leonidas},
  journal={ACM Transactions on Graphics (TOG)},
  volume={35},
  number={6},
  pages={1--12},
  year={2016},
  publisher={ACM New York, NY, USA}
}

@inproceedings{qi2017pointnet,
  title={Pointnet: Deep learning on point sets for 3d classification and segmentation},
  author={Qi, Charles R and Su, Hao and Mo, Kaichun and Guibas, Leonidas J},
  booktitle={Proceedings of the IEEE/CVF Conference on Computer Vision and Pattern Recognition (CVPR)},
  pages={652--660},
  year={2017}
}

@article{wang2019DGCNN,
  title={Dynamic graph cnn for learning on point clouds},
  author={Wang, Yue and Sun, Yongbin and Liu, Ziwei and Sarma, Sanjay E and Bronstein, Michael M and Solomon, Justin M},
  journal={ACM Transactions on Graphics (TOG)},
  volume={38},
  number={5},
  pages={1--12},
  year={2019},
  publisher={Acm New York, NY, USA}
}

@article{qi2017pointnet++,
  title={Pointnet++: Deep hierarchical feature learning on point sets in a metric space},
  author={Qi, Charles Ruizhongtai and Yi, Li and Su, Hao and Guibas, Leonidas J},
  journal={Advances in Neural Information Processing Systems (NeurIPS)},
  volume={30},
  year={2017}
}

@article{qian2022pointnext,
  title={Pointnext: Revisiting pointnet++ with improved training and scaling strategies},
  author={Qian, Guocheng and Li, Yuchen and Peng, Houwen and Mai, Jinjie and Hammoud, Hasan and Elhoseiny, Mohamed and Ghanem, Bernard},
  journal={Advances in Neural Information Processing Systems (NeurIPS)},
  volume={35},
  pages={23192--23204},
  year={2022}
}

@article{guo2020pointcloud,
  title={Deep learning for 3d point clouds: A survey},
  author={Guo, Yulan and Wang, Hanyun and Hu, Qingyong and Liu, Hao and Liu, Li and Bennamoun, Mohammed},
  journal={IEEE Transactions on Pattern Analysis and Machine Intelligence (PAMI)},
  volume={43},
  number={12},
  pages={4338--4364},
  year={2020},
  publisher={IEEE}
}

@article{li2020driving,
  title={Deep learning for lidar point clouds in autonomous driving: A review},
  author={Li, Ying and Ma, Lingfei and Zhong, Zilong and Liu, Fei and Chapman, Michael A and Cao, Dongpu and Li, Jonathan},
  journal={IEEE Transactions on Neural Networks and Learning Systems (TNNLS)},
  volume={32},
  number={8},
  pages={3412--3432},
  year={2020},
  publisher={IEEE}
}

@article{gu2019badnets,
  title={Badnets: Evaluating backdooring attacks on deep neural networks},
  author={Gu, Tianyu and Liu, Kang and Dolan-Gavitt, Brendan and Garg, Siddharth},
  journal={IEEE Access},
  volume={7},
  pages={47230--47244},
  year={2019},
  publisher={IEEE}
}

@inproceedings{feng2022fiba,
  title={FIBA: Frequency-Injection based Backdoor Attack in Medical Image Analysis},
  author={Feng, Yu and Ma, Benteng and Zhang, Jing and Zhao, Shanshan and Xia, Yong and Tao, Dacheng},
  booktitle={Proceedings of the IEEE/CVF Conference on Computer Vision and Pattern Recognition (CVPR)},
  pages={20876--20885},
  year={2022}
}

@article{chen2017blended,
  title={Targeted backdoor attacks on deep learning systems using data poisoning},
  author={Chen, Xinyun and Liu, Chang and Li, Bo and Lu, Kimberly and Song, Dawn},
  journal={arXiv preprint arXiv:1712.05526},
  year={2017}
}

@InProceedings{Bai_2024_BadClipOp,
    author    = {Bai, Jiawang and Gao, Kuofeng and Min, Shaobo and Xia, Shu-Tao and Li, Zhifeng and Liu, Wei},
    title     = {BadCLIP: Trigger-Aware Prompt Learning for Backdoor Attacks on CLIP},
    booktitle = {Proceedings of the IEEE/CVF Conference on Computer Vision and Pattern Recognition (CVPR)},
    month     = {June},
    year      = {2024},
    pages     = {24239-24250}
}

@inproceedings{li2021pointba,
  title={Pointba: Towards backdoor attacks in 3d point cloud},
  author={Li, Xinke and Chen, Zhirui and Zhao, Yue and Tong, Zekun and Zhao, Yabang and Lim, Andrew and Zhou, Joey Tianyi},
  booktitle={Proceedings of the IEEE/CVF International Conference on Computer Vision (ICCV)},
  pages={16492--16501},
  year={2021}
}

@inproceedings{fan2024IBAPC,
  title={Invisible backdoor attack against 3D point cloud classifier in graph spectral domain},
  author={Fan, Linkun and He, Fazhi and Si, Tongzhen and Tang, Wei and Li, Bing},
  booktitle={Proceedings of the AAAI Conference on Artificial Intelligence (AAAI)},
  volume={38},
  number={19},
  pages={21072--21080},
  year={2024}
}

@article{gao2023IRBA,
  title={Imperceptible and robust backdoor attack in 3d point cloud},
  author={Gao, Kuofeng and Bai, Jiawang and Wu, Baoyuan and Ya, Mengxi and Xia, Shu-Tao},
  journal={IEEE Transactions on Information Forensics and Security (TIFS)},
  volume={19},
  pages={1267--1282},
  year={2023},
  publisher={IEEE}
}

@inproceedings{hu2022exploringdevil,
  title={Exploring the devil in graph spectral domain for 3d point cloud attacks},
  author={Hu, Qianjiang and Liu, Daizong and Hu, Wei},
  booktitle={European Conference on Computer Vision (ECCV)},
  pages={229--248},
  year={2022},
  organization={Springer}
}

@inproceedings{lou2024hide,
  title={Hide in thicket: Generating imperceptible and rational adversarial perturbations on 3d point clouds},
  author={Lou, Tianrui and Jia, Xiaojun and Gu, Jindong and Liu, Li and Liang, Siyuan and He, Bangyan and Cao, Xiaochun},
  booktitle={Proceedings of the IEEE/CVF Conference on Computer Vision and Pattern Recognition (CVPR)},
  pages={24326--24335},
  year={2024}
}

@InProceedings{Yang_2024_CVPR,
    author    = {Yang, Sheng and Bai, Jiawang and Gao, Kuofeng and Yang, Yong and Li, Yiming and Xia, Shu-Tao},
    title     = {Not All Prompts Are Secure: A Switchable Backdoor Attack Against Pre-trained Vision Transfomers},
    booktitle = {Proceedings of the IEEE/CVF Conference on Computer Vision and Pattern Recognition (CVPR)},
    month     = {June},
    year      = {2024},
    pages     = {24431-24441}
}

@inproceedings{xiang2021backdoor,
  title={A backdoor attack against 3d point cloud classifiers},
  author={Xiang, Zhen and Miller, David J and Chen, Siheng and Li, Xi and Kesidis, George},
  booktitle={Proceedings of the IEEE/CVF International Conference on Computer Vision (ICCV)},
  pages={7597--7607},
  year={2021}
}

@article{zhang2022pointmae,
  title={Point-m2ae: multi-scale masked autoencoders for hierarchical point cloud pre-training},
  author={Zhang, Renrui and Guo, Ziyu and Gao, Peng and Fang, Rongyao and Zhao, Bin and Wang, Dong and Qiao, Yu and Li, Hongsheng},
  journal={Advances in Neural Information Processing Systems (NeurIPS)},
  volume={35},
  pages={27061--27074},
  year={2022}
}

@article{liu2023meet,
  title={Point cloud attacks in graph spectral domain: When 3d geometry meets graph signal processing},
  author={Liu, Daizong and Hu, Wei and Li, Xin},
  journal={IEEE Transactions on Pattern Analysis and Machine Intelligence (PAMI)},
  volume={46},
  number={5},
  pages={3079--3095},
  year={2023},
  publisher={IEEE}
}

@inproceedings{zhou2019dup,
  title={Dup-net: Denoiser and upsampler network for 3d adversarial point clouds defense},
  author={Zhou, Hang and Chen, Kejiang and Zhang, Weiming and Fang, Han and Zhou, Wenbo and Yu, Nenghai},
  booktitle={Proceedings of the IEEE/CVF International Conference on Computer Vision (ICCV)},
  pages={1961--1970},
  year={2019}
}

@inproceedings{zheng2019saliencymap,
  title={Pointcloud saliency maps},
  author={Zheng, Tianhang and Chen, Changyou and Yuan, Junsong and Li, Bo and Ren, Kui},
  booktitle={Proceedings of the IEEE/CVF International Conference on Computer Vision (ICCV)},
  pages={1598--1606},
  year={2019}
}

@inproceedings{saha2022backdoor,
  title={Backdoor attacks on self-supervised learning},
  author={Saha, Aniruddha and Tejankar, Ajinkya and Koohpayegani, Soroush Abbasi and Pirsiavash, Hamed},
  booktitle={Proceedings of the IEEE/CVF Conference on Computer Vision and Pattern Recognition (CVPR)},
  pages={13337--13346},
  year={2022}
}

@inproceedings{yuan2023transformer,
  title={You are catching my attention: Are vision transformers bad learners under backdoor attacks?},
  author={Yuan, Zenghui and Zhou, Pan and Zou, Kai and Cheng, Yu},
  booktitle={Proceedings of the IEEE/CVF Conference on Computer Vision and Pattern Recognition (CVPR)},
  pages={24605--24615},
  year={2023}
}

@article{gao2024backdoor,
  title={Backdoor attack with sparse and invisible trigger},
  author={Gao, Yinghua and Li, Yiming and Gong, Xueluan and Li, Zhifeng and Xia, Shu-Tao and Wang, Qian},
  journal={IEEE Transactions on Information Forensics and Security (TIFS)},
  year={2024},
  publisher={IEEE}
}

@inproceedings{selvaraju2017gradcam,
  title={Grad-cam: Visual explanations from deep networks via gradient-based localization},
  author={Selvaraju, Ramprasaath R and Cogswell, Michael and Das, Abhishek and Vedantam, Ramakrishna and Parikh, Devi and Batra, Dhruv},
  booktitle={Proceedings of the IEEE/CVF International Conference on Computer Vision (ICCV)},
  pages={618--626},
  year={2017}
}

@article{li2022untargeted,
  title={Untargeted backdoor watermark: Towards harmless and stealthy dataset copyright protection},
  author={Li, Yiming and Bai, Yang and Jiang, Yong and Yang, Yong and Xia, Shu-Tao and Li, Bo},
  journal={Advances in Neural Information Processing Systems (NeurIPS)},
  volume={35},
  pages={13238--13250},
  year={2022}
}

@article{bian2024iba,
  title={iba: Backdoor attack on 3d point cloud via reconstructing itself},
  author={Bian, Yuhao and Tian, Shengjing and Liu, Xiuping},
  journal={IEEE Transactions on Information Forensics and Security (TIFS)},
  year={2024},
  publisher={IEEE}
}

@article{yu20213d,
  title={3d medical point transformer: Introducing convolution to attention networks for medical point cloud analysis},
  author={Yu, Jianhui and Zhang, Chaoyi and Wang, Heng and Zhang, Dingxin and Song, Yang and Xiang, Tiange and Liu, Dongnan and Cai, Weidong},
  journal={arXiv preprint arXiv:2112.04863},
  year={2021}
}

@article{zhu2025point,
  title={Point cloud matters: Rethinking the impact of different observation spaces on robot learning},
  author={Zhu, Haoyi and Wang, Yating and Huang, Di and Ye, Weicai and Ouyang, Wanli and He, Tong},
  journal={Advances in Neural Information Processing Systems (NeurIPS)},
  volume={37},
  pages={77799--77830},
  year={2025}
}

@inproceedings{hu2023pointcrt,
  title={Pointcrt: Detecting backdoor in 3d point cloud via corruption robustness},
  author={Hu, Shengshan and Liu, Wei and Li, Minghui and Zhang, Yechao and Liu, Xiaogeng and Wang, Xianlong and Zhang, Leo Yu and Hou, Junhui},
  booktitle={Proceedings of the ACM International Conference on Multimedia (MM)},
  pages={666--675},
  year={2023}
}

@article{lian2025cross,
  title={Cross-Modal Driven Object Restoration for 3D Point Cloud Backdoor Defense},
  author={Lian, Jiawei and Du, Xia and Liu, Jianghua and Hui, Le and Yang, Jian},
  journal={IEEE Transactions on Information Forensics and Security},
  year={2025},
  publisher={IEEE}
}

@inproceedings{xiang2022detecting,
  title={Detecting backdoor attacks against point cloud classifiers},
  author={Xiang, Zhen and Miller, David J and Chen, Siheng and Li, Xi and Kesidis, George},
  booktitle={ICASSP 2022-2022 IEEE International Conference on Acoustics, Speech and Signal Processing (ICASSP)},
  pages={3159--3163},
  year={2022},
  organization={IEEE}
}

@article{hou2025flare,
  title={Flare: Towards universal dataset purification against backdoor attacks},
  author={Hou, Linshan and Luo, Wei and Hua, Zhongyun and Chen, Songhua and Zhang, Leo Yu and Li, Yiming},
  journal={IEEE Transactions on Information Forensics and Security},
  year={2025},
  publisher={IEEE}
}

@article{song2026wpda,
  title={{WPDA}: Frequency-based backdoor attack with wavelet packet decomposition},
  author={Song, Zhengyao and Li, Yongqiang and Yuan, Danni and Liu, Li and Wei, Shaokui and Wu, Baoyuan},
  journal={Neural Networks},
  volume={194},
  pages={108074},
  year={2026},
  doi={10.1016/j.neunet.2025.108074}
}

@article{yang2026featuretrojan,
  title={{FeatureTrojan}: Boosting stealthy and steady backdoor attacks with feature poisoning and fine-tuning injection},
  author={Yang, Rui and Sun, Qindong and Cao, Han and Lin, Kai and Shen, Chao},
  journal={Neural Networks},
  volume={198},
  pages={108717},
  year={2026},
  doi={10.1016/j.neunet.2026.108717}
}

@article{li2025dtgba,
  title={{DTGBA}: A stronger graph backdoor attack with dual triggers},
  author={Li, Ding and Xia, Hui and Li, Xin and Zhang, Rui and Ma, Mingda},
  journal={Neural Networks},
  volume={190},
  pages={107726},
  year={2025},
  doi={10.1016/j.neunet.2025.107726}
}

@article{huang2026mfbd,
  title={MFBD: Model-free backdoor defense based on vision-language pre-trained models},
  author={Huang, Rui and Hao, Mengjia and Wang, Hechuan and Xing, Yan and Zhang, Yuxiang},
  journal={Neural Networks},
  pages={108772},
  year={2026},
  publisher={Elsevier}
}

@article{peng2025backdoor,
  title={Backdoor samples detection based on perturbation discrepancy consistency in pre-trained language models},
  author={Peng, Zuquan and Fu, Jianming and Zou, Lixin and Zheng, Li and Ren, Yanzhen and Peng, Guojun},
  journal={Neural Networks},
  pages={108025},
  year={2025},
  publisher={Elsevier}
}
}

\end{document}